\documentclass{article}
\usepackage{spconf,amsmath,epsfig, url,microtype, graphicx, booktabs, multirow}
\usepackage[table,xcdraw]{xcolor}
\title{ENHANCING DEEP LEARNING PERFORMANCE ON BURNED AREA DELINEATION \\ FROM SPOT-6/7 IMAGERY FOR EMERGENCY MANAGEMENT}
\name{María Rodríguez$^{1,2}$, Minh-Tan Pham$^{1}$, Martin Sudmanns$^{2}$, Quentin Poterek$^{3}$, Oscar Narvaez$^{3}$}
\address{$^{1}$ IRISA, Université Bretagne Sud, UMR 6074, Vannes, France\\
         $^{2}$ University of Salzburg, Department of Geoinformatics, Salzburg, Austria \\
         $^{3}$ ICube-SERTIT, Université de Strasbourg, Illkirch Graffenstaden, France}
  
\begin{document}
%
\maketitle
\begin{abstract}
After a wildfire, delineating burned areas (BAs) is crucial for quantifying damages and supporting ecosystem recovery. Current BA mapping approaches rely on computer vision models trained on post-event remote sensing imagery, but often overlook their applicability to time-constrained emergency management scenarios. This study introduces a supervised semantic segmentation workflow aimed at boosting both the performance and efficiency of BA delineation. It targets SPOT-6/7 imagery due to its very high resolution 
and on-demand availability. Experiments are evaluated based on Dice score, Intersection over Union, and inference time. The results show that U-Net 
and SegFormer 
models perform similarly with limited training data. However, SegFormer requires more resources, challenging its practical use in emergencies. Incorporating land cover 
data as an auxiliary task enhances model robustness without increasing inference time. Lastly, Test-Time Augmentation 
improves BA delineation performance but raises inference time, which can be mitigated with optimization methods like Mixed Precision.
\end{abstract}
\begin{keywords}
remote sensing, burned area delineation, emergency management, test-time augmentation, mixed precision.
\end{keywords}
\section{Introduction}
\label{sec:intro}
Wildfires considerable impact Earth's dynamics across various scales. They release greenhouse gases and absorptive aerosols, contributing to climate change. Additionally, wildfires affect ecosystems and have socio-economic consequences, including threats to human health from smoke inhalation and property damage. After wildfires, authorities require the extent of the burned areas (BAs) to estimate the impact on wildlife habitats and human settlements, quantify damages, and develop restoration strategies~\cite{mcentire2021disaster}. 

In emergency management, accurate BA delineation relies on appropriate data and robust methods. These methods should distinguish BA from other types of land cover (LC) changes, such as seasonal soil moisture variations or harvesting and account for the spectral, spatial, and temporal variability caused by different fire severities and LC types~\cite{randerson2012global}. Moreover, the delineation must be delivered in the shortest timeframe possible to support immediate response efforts after the disaster~\cite{ajmar2015rapid}.

Remote sensing is the primary means for mapping BAs. For instance, the Moderate Resolution Imaging Spectroradiometer (MODIS) offers a global mapping solution but may result in underestimations in local contexts. Higher-resolution imagery from Sentinel-2 has considerably improved delineation accuracy and facilitated monitoring. Still, Very High Resolution (VHR) sensors like PlanetScope and SPOT are usually preferred for emergency management due to their high detail, semantic content, and on-demand availability~\cite{ghali2023deep,narvaez2023burnt}.

Multiple approaches to map BAs based on satellite imagery exist. Traditional methods involve thresholding the difference between pre-and post-event vegetation indices~\cite{kasischke1993monitoring}, but they depend on the spatial setting and require extensive post-processing. Machine learning algorithms like Random Forest have also been explored for this purpose~\cite{gibson2020remote}. While effective, they often exhibit long computational times when applied to VHR imagery. Recently, deep learning (DL) semantic segmentation methods using Convolutional Neural Networks (CNNs) like U-Net and Transformer-based architectures like SegFormer have shown promising results by considering the contextual information of post-wildfire imagery \cite{arnaudo2023robust,han2024burned}.

Previous approaches focused on improving BA segmentation accuracy, often neglecting their pertinence in time-critical emergency management contexts. Thus, this paper aims to develop a supervised semantic segmentation workflow for delineating BAs using SPOT-6/7 imagery, optimizing both performance and inference speed. Specifically, this study evaluates: 1) CNN-based U-Net against Transformer-based SegFormer architectures; 2) Single Task Learning (STL) framework versus Multi-Task Learning~(MTL) incorporating an auxiliary LC segmentation head; 3)~Techniques such as Test-Time Augmentation (TTA)~\cite{moshkov2020test} and Mixed Precision (MP)~\cite{liu2021post}.

\section{Dataset}
\label{sec:dataset}
\subsection{Study Area}
\label{ssec:studyarea}
The study areas are ten wildfire events with the Copernicus Emergency Management Service (CEMS) activation between 2016 and 2024. Nine events in Greece are utilized to train, validate, and test the models; One in Spain is used to assess the models' generalization to new, unseen scenarios.

The Greek events are spread across the country and are selected to capture a variety of environmental conditions, typical of a Mediterranean climate with mild, rainy winters and hot, dry summers. The main LC types in Greece include tree cover, grassland, and shrubland~\cite{katagis2022assessing}. In Spain, the event occurred near Navafría (Segovia), which also has a Mediterranean climate. Unlike the Greek events, the primary LC in this region is grassland, nearly three times more prevalent than tree cover.

\subsection{Data}
\label{ssec:data}
For each wildfire event, five datasets are used, primarily sourced from CEMS Rapid Mapping Portfolio. They include the Area of Interest (AOI), a cloud mask, the BA delineation, and the post-wildfire SPOT-6/7 image. This image is pansharpened to 1.5 m, has four bands (Blue, Green, Red, and NIR), and is radiometrically corrected and orthorectified. In all AOIs there is an imbalance with more non-burned than burned pixels. The fifth dataset is the European Space Agency~(ESA) WorldCover 2021, consisting of 10 m LC maps. 

During data preparation, the BA delineations and cloud masks are converted into binary rasters and co-registered with the corresponding SPOT-6/7 imagery. The cloud mask is subtracted from the BA delineation to prevent misclassification of clouds as BAs. Finally, the LC maps are resampled to 1.5~m using nearest neighbor interpolation to match the SPOT-6/7 imagery's resolution, with clouds added as an extra category. 

The resulting rasters are clipped to the AOIs and divided into smaller patches of 512x512 pixels. For Greece, patches are created without overlap. In contrast, for Spain, a 20\% overlap is included to aid in final map mosaicking during prediction. After data preparation, Greece and Spain have 5160 and 90 patches, respectively.

\section{Methodology}
\label{sec:methodology}
\subsection{Frameworks}
\label{ssec:frameworks}
Building on the findings of the previous research~\cite{arnaudo2023robust}, two modelling frameworks are configured: STL and MTL. STL exclusively focuses on delineating BAs, whereas MTL also incorporates an auxiliary task of LC segmentation.

For STL training, post-event satellite image patches~\( x_i \) are paired with their corresponding ground truth BA delineations~\( y_{BA_i} \). Specifically, \( x_i \) pass through a model with an encoder-decoder architecture to produce feature representations. A single segmentation head~\( h_{BA} \) then generates predicted BA probability maps~\( \hat{y}_{BA} \). The training process minimizes the loss function~\( L_{STL} \), equivalent to the BA delineation task loss~\( L_{BA} \):
\[L_{BA} = L(y_{BA}, \hat{y}_{BA}) \tag{1}\]

\begin{figure}[ht]
    \centering
    \includegraphics[width=\linewidth]{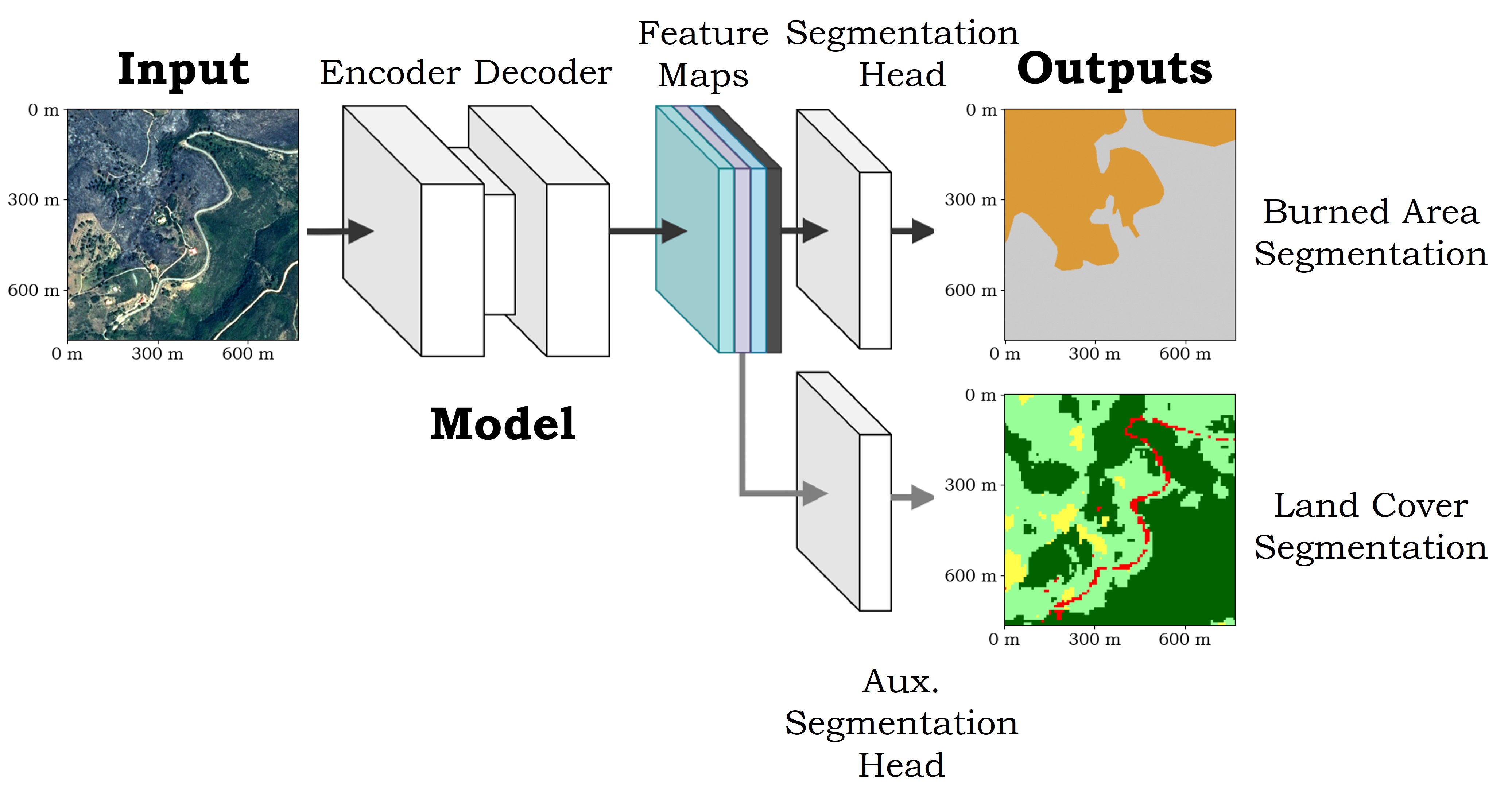}
    \caption{MTL framework for BA delineation, consisting of a model encoder-decoder model and two segmentation heads. Modified from~\cite{arnaudo2023robust}.}
    \label{fig:fig1}
\end{figure}

MTL has the same encoder-decoder model architecture as STL to produce feature maps~(Fig.~\ref{fig:fig1}). However, these features are shared between two segmentation heads: \( h_{BA} \) for BA probability maps~\( \hat{y}_{BA} \), and \( h_{LC} \) for LC probability maps~\( \hat{y}_{LC} \). The training process minimizes a loss function~\( L_{MTL} \), which is the weighted sum of the losses for both tasks \( L_{BA} \) and \( L_{LC} \), with \( \lambda \) a weighting factor:
\[L_{MTL} = L_{BA} + \lambda L_{LC} \tag{2}\]

\subsection{Models}
\label{ssec:models}
Two architectures are explored: a CNN-based U-Net with a ResNet34 backbone (UNet-RN34, 24.4M parameters) and a Transformer-based SegFormer with an MiT-B2 backbone (SegFormer-B2, 27.4M parameters). Both models are selected to have comparable parameter counts, to ensure performance differences are attributable to architectural design and principles.

\subsection{Metrics}
\label{ssec:metrics}
The performance of the models is evaluated using Dice score~(Dice) and Intersection over Union (IoU), which are chosen to address the imbalance in class distribution. Additionally, inference time is measured to assess real-world applicability in rapid mapping scenarios. 

\subsection{Test-Time Augmentation (TTA)}
\label{ssec:tta}
TTA~\cite{moshkov2020test} is a technique designed to boost model performance by using augmented versions of the testing samples during inference. The process consists of four steps for each test sample: 1) apply geometric augmentations such as horizontal and vertical flips, along with rotations; 2) predict on both the original and augmented samples; 3) revert the augmentations on the output probability maps; 4) average the probability maps to produce the final model output.

\subsection{Mixed Precision (MP)}
\label{ssec:mp}
MP~\cite{liu2021post} is used to improve the efficiency of DL models. It combines single precision (32-bit floating-point, FP32) with half-precision (16-bit floating-point, FP16). The primary advantage is that FP16 reduces memory bandwidth, enabling GPUs to process and transfer data faster without compromising accuracy compared to full precision~\cite{liu2021post}. Implementing MP requires two main changes: using FP16 where suitable in the model and adding loss scaling to preserve the integrity of small gradient values. The first change assigns the optimal data type to each operation, e.g., convolutions run faster in FP16, while batch normalization utilizes FP32 for a broader dynamic range~\cite{automatic2024}. The second change prevents FP16 gradients from underflowing during backpropagation.

\section{Experimental Settings}
\label{sec:settings}
For Greece, patches are split block-wise, as proposed by~\cite{gella2023unsupervised}. Each AOI is divided into 2 km x 2 km blocks and randomly split into training (80\%), validation (20\%), and testing (10\%) sets along with their contained patches. This method ensures independent sets with similar burned-non-burned pixel distributions. For Spain, all patches form the prediction set without subdivision. 

STL with UNet-RN34 is selected as a baseline, in-line with prior studies~\cite{arnaudo2023robust,colomba2022dataset}. Experiments are initially conducted over 20 epochs to assess the impact of various configurations, including Transformer-based models like SegFormer-B2 and the addition of a LC segmentation task for BA delineation. Techniques such as TTA and MP are also investigated. The final models are trained with the most beneficial configurations and techniques over an extended 40 epochs.

Experiments are conducted on an NVIDIA RTX A6000 GPU, using PyTorch Lightning as the DL framework. The code is available here\footnote{\url{https://github.com/mariarodriguezn/BurnedAreasDelineation}}. Key hyperparameters include a batch size of 8, an AdamW optimizer, learning rates of \( 1 \times 10^{-4} \) for UNet-RN34 and \( 6 \times 10^{-5} \) for SegFormer-B2, and a weight decay of \( 1 \times 10^{-4} \). Dice loss is used for both BA and LC tasks, with \( \lambda \) calibrated to 0.3 and 0.2 for UNet-RN34 and SegFormer-B2, respectively. Training data augmentations integrate random horizontal and vertical flips and 90° rotations, with a probability of 0.5. All models are initialized with ImageNet weights.

\section{Results and Discussion}
\label{sec:results}
\subsection{Frameworks and Models Evaluation}
\label{ssec:evaluation}
The results align with the findings reported by~\cite{arnaudo2023robust} based on Sentinel-2 data. MTL consistently outperforms STL in Dice and IoU metrics (Table~\ref{tab1}). This is attributed to the additional LC task, providing extra contextual information and regularizing the BA learning process. Although MTL increases the number of trainable parameters, adding approximately 5 minutes per epoch during training, inference times remain unchanged compared to STL. This is because the auxiliary head is only used for training and dropped during inference.

\begin{table}[ht]
\centering
\caption{Experimental results of STL and MTL with SegFormer-B2 and UNet-RN34 in Greece. The highlighted row indicates the baseline results.}
\label{tab1}
\vspace{0.3em}
\resizebox{\columnwidth}{!}{%
\begin{tabular}{@{}cccccc@{}}
\toprule
\rowcolor[HTML]{FFFFFF} 
\textbf{Framework} & \textbf{Model} & \textbf{Dice} & \textbf{IoU} & \textbf{\begin{tabular}[c]{@{}c@{}}Train Time \\ (per epoch)\\ (min)\end{tabular}} & \textbf{\begin{tabular}[c]{@{}c@{}}Inference\\ Time\\ (min)\end{tabular}} \\ \midrule
\rowcolor[HTML]{FFFFFF} 
\cellcolor[HTML]{FFFFFF} & \textbf{UNet-RN34} & 0.9065 & 0.8806 & \ 9.6887 & \ 1.4259 \\
\rowcolor[HTML]{FFFFFF} 
\multirow{-2}{*}{\cellcolor[HTML]{FFFFFF}\textbf{STL}} & \textbf{Segformer-B2} & \ 0.9136 & \ 0.8876 & 24.1893 & 1.4716 \\ \midrule
\rowcolor[HTML]{FFFFFF} 
\cellcolor[HTML]{FFFFFF} & \textbf{UNet-RN34} & \ 0.9314 & \ 0.9075 & \ 14.7623 & \ 1.4223 \\
\rowcolor[HTML]{FFFFFF} 
\multirow{-2}{*}{\cellcolor[HTML]{FFFFFF}\textbf{MTL}} & \textbf{Segformer-B2} & 0.9299 & 0.9065 & 29.6289 & 1.4792 \\ \bottomrule
\end{tabular}%
}
\end{table}

\begin{figure}[ht]
    \centering
    \includegraphics[width=\linewidth]{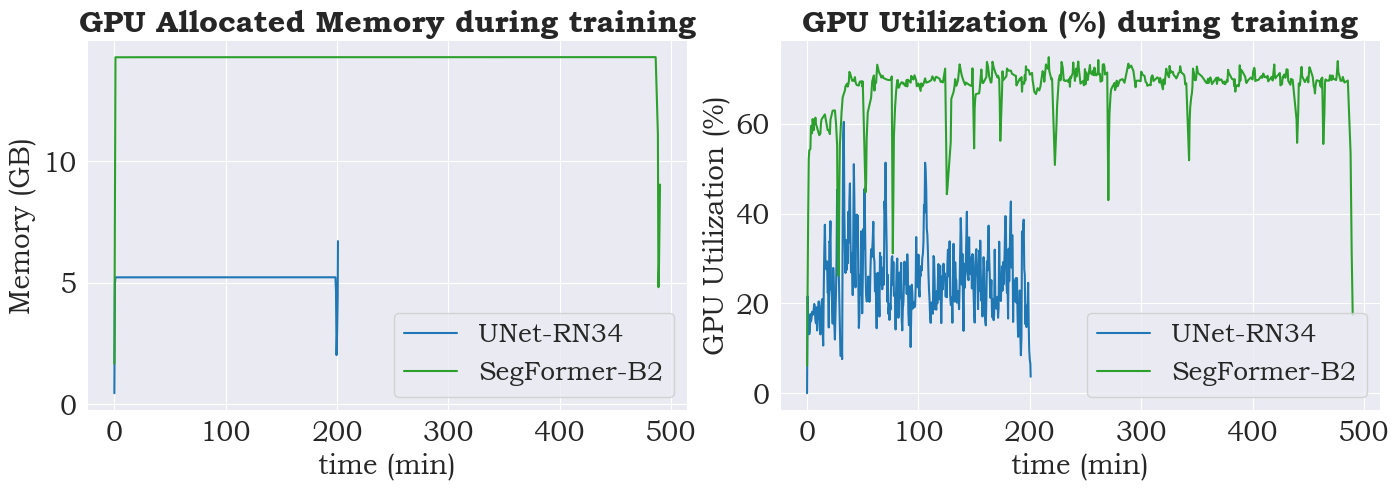}
    \caption{Allocated GPU memory (GB) and utilization (\%) during training for MTL with Unet-RN34 and SegFormer-B2 in Greece.}
    \label{fig:fig2}
\end{figure}

SegFormer-B2 and UNet-RN34 achieve similar Dice and IoU scores. However, SegFormer-B2 extends training time by an additional 15 minutes per epoch, demands approximately 2.4 times the GPU memory, and consistently records higher GPU utilization per epoch~(Fig.~\ref{fig:fig2}). The higher consumption is due to its attention layers, where computational load scales quadratically with the number of tokens involved. Furthermore, SegFormer-B2 requires interpolation to upscale its outputs, which are one-quarter of the original dimensions. Despite SegFormer’s inherent architectural optimizations to mitigate the computational burden, U-Net remains more resource-efficient while delivering similar performance metrics in scenarios with limited training data.

Given these results, subsequent techniques evaluated to refine the BA segmentation task for STL and MTL use only the UNet-RN34 model.

\subsection{Effect of training and inference techniques}
\label{ssec:effects}
TTA improve the BA segmentation by increasing Dice and IoU metrics for both frameworks (Table~\ref{tab2}). Still, a primary drawback of TTA is the increased inference time (ca. 60\%).

\begin{table}[ht]
\centering
\caption{Impact of TTA on BAs segmentation metrics for STL and MTL with UNet-RN34 in Greece.}
\label{tab2}
\vspace{0.3em}
\resizebox{\columnwidth}{!}{%
\begin{tabular}{@{}
>{\columncolor[HTML]{FFFFFF}}c 
>{\columncolor[HTML]{FFFFFF}}c 
>{\columncolor[HTML]{FFFFFF}}c 
>{\columncolor[HTML]{FFFFFF}}c 
>{\columncolor[HTML]{FFFFFF}}c 
>{\columncolor[HTML]{FFFFFF}}c @{}}
\toprule
\textbf{Framework} & \textbf{Model} & \textbf{Technique} & \textbf{Dice} & \textbf{IoU} & \textbf{\begin{tabular}[c]{@{}c@{}}Inference\\      Time (min)\end{tabular}} \\ \midrule
\cellcolor[HTML]{FFFFFF} & \cellcolor[HTML]{FFFFFF} & \textbf{No TTA} & 0.9065 & 0.8806 & \ 1.4259 \\
\multirow{-2}{*}{\cellcolor[HTML]{FFFFFF}\textbf{STL}} & \cellcolor[HTML]{FFFFFF} & \textbf{TTA} & \ 0.9324 & \ 0.9078 & 2.2613 \\ \cmidrule(r){1-1} \cmidrule(l){3-6} 
\cellcolor[HTML]{FFFFFF} & \cellcolor[HTML]{FFFFFF} & \textbf{No TTA} & 0.9314 & 0.9075 & \ 1.4223 \\
\multirow{-2}{*}{\cellcolor[HTML]{FFFFFF}\textbf{MTL}} & \multirow{-4}{*}{\cellcolor[HTML]{FFFFFF}\textbf{UNet-RN34}} & \textbf{TTA} & \ 0.9349 & \ 0.9116 & 2.2789 \\ \bottomrule
\end{tabular}%
}
\end{table}

Implementing MP during training and inference reduces training time per epoch by 30\% and inference time by 51\% without scarifying performance (Table~\ref{tab3}). 

\begin{table}[ht]
\centering
\caption{Comparative results of including MP during training and inference for STL and MTL with UNet-RN34 in Greece.}
\label{tab3}
\vspace{0.3em}
\resizebox{\columnwidth}{!}{%
\begin{tabular}{@{}
>{\columncolor[HTML]{FFFFFF}}c 
>{\columncolor[HTML]{FFFFFF}}c 
>{\columncolor[HTML]{FFFFFF}}c 
>{\columncolor[HTML]{FFFFFF}}c 
>{\columncolor[HTML]{FFFFFF}}c 
>{\columncolor[HTML]{FFFFFF}}c 
>{\columncolor[HTML]{FFFFFF}}c @{}}
\toprule
\textbf{Framework} & \textbf{Model} & \textbf{Technique} & \textbf{Dice} & \textbf{IoU} & \textbf{\begin{tabular}[c]{@{}c@{}}Training Time\\      (per epoch)\\      (min)\end{tabular}} & \textbf{\begin{tabular}[c]{@{}c@{}}Inference\\      Time\\      (min)\end{tabular}} \\ \midrule
\cellcolor[HTML]{FFFFFF} & \cellcolor[HTML]{FFFFFF} & \textbf{No MP} & 0.9065 & 0.8806 & 9.6887 & 1.4259 \\
\multirow{-2}{*}{\cellcolor[HTML]{FFFFFF}\textbf{STL}} & \cellcolor[HTML]{FFFFFF} & \textbf{MP} & 0.9198 & 0.8951 & 6.6590 & 0.6900 \\ \cmidrule(r){1-1} \cmidrule(l){3-7} 
\cellcolor[HTML]{FFFFFF} & \cellcolor[HTML]{FFFFFF} & \textbf{No MP} & 0.9314 & 0.9075 & 14.7623 & 1.4223 \\
\multirow{-2}{*}{\cellcolor[HTML]{FFFFFF}\textbf{MTL}} & \multirow{-4}{*}{\cellcolor[HTML]{FFFFFF}\textbf{UNet-RN34}} & \textbf{MP} & 0.9405 & 0.916 & 9.1103 & 0.6893 \\ \bottomrule
\end{tabular}%
}
\end{table}

\subsection{Final Models}
\label{ssec:finalmodels}
The final models, which combine TTA and MP alongside an increased training period, display significant improvements over the baseline for both frameworks in Greece. STL sees a 3.99\% increase in Dice and 4.42\% in IoU, MTL a 4.24\% increase in Dice and 5.35\% in IoU. Both frameworks reduce inference time by 31\%, compared to the baseline.

\subsection{Prediction on Spain with Final Models}
\label{ssec:prediction}
\begin{figure}[ht]
    \centering
    \includegraphics[width=\linewidth]{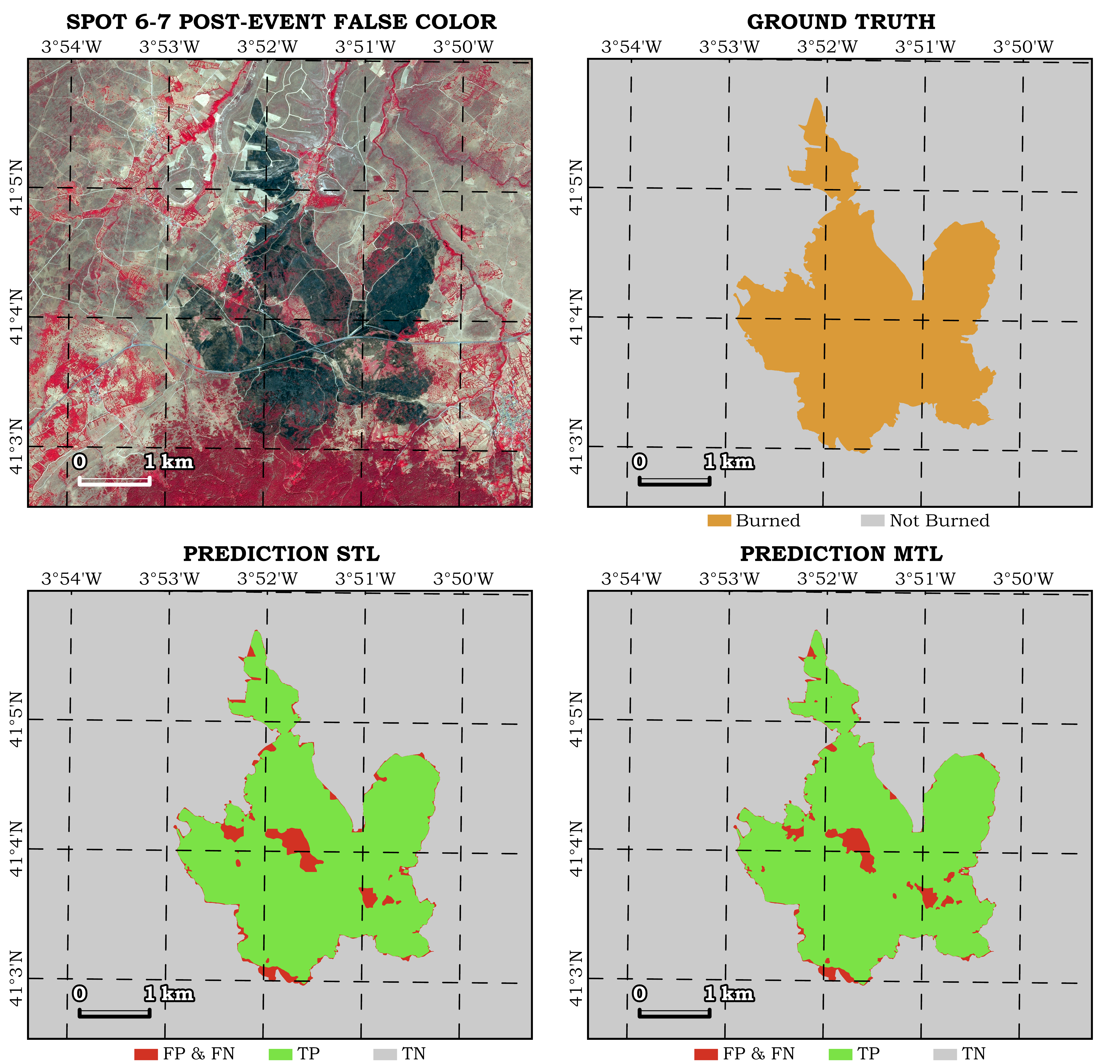}
    \caption{Prediction outputs for Spain.}
    \label{fig:fig3}
\end{figure}

Quantitatively, MTL slightly outperforms STL and the inference times are under one minute~(Table~\ref{tab4}). Qualitatively, the generated maps by both frameworks show similar delineations~(Fig.~\ref{fig:fig3}), with most errors being false negatives~(FN), and occurring in the same areas. A closer inspection of these areas in the post-event imagery indicates that some regions labelled as burned in the ground truth are actually not burned. This discrepancy is likely due to simplifications made by the production site. As anticipated in a supervised learning setting, the results depend on the quality of the annotations.

\begin{table}[ht]
\centering
\caption{Prediction metrics for Spain.}
\label{tab4}
\vspace{0.3em}
\resizebox{\columnwidth}{!}{%
\begin{tabular}{@{}>{\columncolor[HTML]{FFFFFF}}c 
>{\columncolor[HTML]{FFFFFF}}c 
>{\columncolor[HTML]{FFFFFF}}c 
>{\columncolor[HTML]{FFFFFF}}c 
>{\columncolor[HTML]{FFFFFF}}c 
>{\columncolor[HTML]{FFFFFF}}l @{}}
\toprule
\textbf{Framework} & \textbf{Model} & \textbf{Dice} & \textbf{IoU} & \multicolumn{2}{c}{\cellcolor[HTML]{FFFFFF}\textbf{\begin{tabular}[c]{@{}c@{}}Inference\\ Time (min)\end{tabular}}} \\ \midrule
\textbf{STL} & \cellcolor[HTML]{FFFFFF} & 0.8811 & 0.8542 & \multicolumn{2}{c}{\cellcolor[HTML]{FFFFFF}0.1944} \\ \cmidrule(r){1-1} \cmidrule(l){3-6} 
\textbf{MTL} & \multirow{-2}{*}{\cellcolor[HTML]{FFFFFF}\textbf{UNet-RN34}} & 0.9016 & 0.8824 & \multicolumn{2}{c}{\cellcolor[HTML]{FFFFFF}0.2114} \\ \bottomrule
\end{tabular}%
}
\end{table}

\section{Conclusion}
\label{sec:conclusion}

This paper presents a supervised semantic segmentation workflow to delineate BAs using SPOT-6/7 imagery, improving both performance and efficiency in emergency management contexts. It evaluates segmentation architectures, modelling frameworks, and training techniques. With comparable parameter counts U-Net and SegFormer perform similarly but SegFormer requires more computational resources. Incorporating a multitask learning approach by adding LC segmentation as an auxiliary task enhances robustness and generalization without extending inference times. TTA improves the metrics by averaging predictions from multiple augmented versions of the test data, thereby reducing the impact of individual errors. Combining TTA with MP is beneficial in emergency contexts, as it increases efficiency without compromising performance.

Limitations of supervised learning exist, e.g., the dependency on label quality and data distribution. Models may underperform if training and test scenarios differ. Future research should investigate recent MTL \cite{le2023data}, self-supervised learning \cite{berg2022self} and domain adaptation techniques \cite{tuia2016domain} to enhance generalization across geographic regions.

\bibliographystyle{ieeetr}
\bibliography{refs}

\end{document}